# Real-time Traffic Flow Parameters Prediction with Basic Safety Messages at Low Penetration of Connected Vehicles

Mizanur Rahman, Mashrur Chowdhury, *Senior Member, IEEE,* and Jerome McClendon

*Abstract*— The expected low market penetration of connected vehicles (CVs) in the near future could be a constraint in estimating traffic flow parameters, such as average travel speed of a roadway segment and average space-headway between vehicles from the CV broadcasted data. This estimated traffic flow parameters from low penetration of connected vehicles become noisy compared to 100% penetration of CVs, and such noise reduces the real-time prediction accuracy of a machine-learning model, such as the accuracy of long short-term memory (LSTM) model in terms of predicting traffic flow parameters. The accurate prediction of the parameters is important for future traffic condition assessment. To improve the prediction accuracy using noisy traffic flow parameters, which is constrained by limited CV market penetration and limited CV data, we developed a real-time traffic-data prediction model that combines LSTM with Kalman filter based Rauch–Tung–Striebel (RTS) noise reduction model. We conducted a case study using the Enhanced Next Generation Simulation (NGSIM) dataset, which contains vehicle trajectory data for every one-tenth of a second, to evaluate the performance of this prediction model. Compared to a baseline LSTM model performance, for only 5% penetration of CVs, the analyses revealed that combined LSTM/RTS model reduced the mean absolute percentage error (MAPE) from 19% to 5% for speed prediction and from 27% to 9% for space-headway prediction. The statistical significance test with a 95% confidence interval confirmed no significant difference in predicted average speed and average space headway using this LSTM/RTS combination with only 5% CV penetration rate.

## I. Introduction

THE evolution of new Intelligent Transportation Systems (ITS) technologies has made possible the Advanced Traffic Management System (ATMS) and Advanced Traveler Information System (ATIS) initiatives to inform travelers about current and future traffic conditions [1]-[3]. These traffic management strategies depend upon the accurate prediction of traffic flow parameters, such as travel speed and space headway between vehicles. They are used for route planning and scheduling to reduce travel time, for future traffic condition assessment, and for energy optimization to reduce fuel consumption [2], [3], [4]-[7].

However, real-time traffic flow parameters prediction is challenging given the dynamic traffic flow on roadways over time [4], [8]-[9]. As such, capturing the temporal relationship over time to predict traffic flow parameters accurately is most important. Although a wide range of inductive loop detectors and video cameras are used to collect this data, the fixed position of these sensors prevents the simultaneous capture of stochastic traffic flow in terms of spatial and temporal variation for a specific roadway segment [4], [10]. Connected Vehicle (CV) technologies that provide interconnection between transportation systems are used to address this difficulty, by allowing vehicles to share Basic Safety Messages (BSMs) via communication both with one another and with transportation roadside infrastructures (e.g., traffic signal, roadside unit) and Traffic Management Centers (TMC). In this CV system, BSMs provide trajectory data (i.e., location, speed, acceleration, and deceleration) of each vehicle for every one-tenth of a second [11], [12]. Providing real-time BSMs with this temporal and spatial variation requires data-driven approaches, such as machine learning models for capturing the non-linearity of traffic patterns.

More specifically, recurrent neural networks (RNNs), a type of machine learning model that captures temporal variation and predict time series data, have been used to predict freeway traffic volume [10]. However, these traditional RNNs, such as Simple RNN and Gated Recurrent Network (GRU), are unable to capture the long temporal dependency in the traffic patterns due to an unexpected traffic event on a roadway, such as a traffic incident that occurs one hour previously may still cause severe congestion in the subsequent two to three hours after the event [4], [10]. Consequently, a special RNN architecture, the Long Short-Term Memory (LSTM) neural network was developed to address these limitations in terms of time series prediction [13].

Over the past decade, the LSTM has been successfully used in robot control, speech recognition, handwriting recognition, human action recognition and univariate traffic prediction via data collected from roadway sensors [4], [10]. However, the near absence of research using LSTM and BSMs from connected vehicles to predict traffic flow parameters limits the near future penetration of connected vehicles, which prevents accurate estimation of current speed or space headway. This inaccuracy in the estimated current speed and space-headway data is referred to as "noise." The noise in the traffic flow parameters can significantly reduce the prediction accuracy of a machine-learning model, such as long short-term memory (LSTM). In addition, a massive amount of data is required to train a deep learning model to accurately predict speed and space headway with noisy traffic flow parameters and to achieve an expected prediction accuracy.

To improve the real-time prediction accuracy with low penetration of CVs, we developed a traffic flow parameters-prediction model that combines the LSTM with a noise reduction model. We first investigated two noise reduction models, the standard Kalman filter and the Kalman filter based Rauch–Tung–Striebel (RTS) data smoothing techniques, to reduce the noise from the traffic flow parameters measured from BSMs. Next, using the resulting filtered data we evaluated the performance of the LSTM prediction model for predicting traffic flow parameters. Using a vehicle penetration rate ranging from 5% to 90%, we used enhanced Next Generation Simulation (NGSIM) data which contain vehicle trajectory data for every one-tenth of a second as the BSMs for the evaluating our LSTM prediction model [14].

The remainder of this paper is broken down into the following sections. Section II describes the contribution of the paper. Section III describes the related work on traffic flow parameters prediction using machine learning, noise reduction models and statistical models for time series prediction, while Section IV presents the traffic flow parameters prediction from BSMs of CVs in a mixed traffic scenario (connected and non-connected vehicles) to explore the impact of noisy data on the prediction of average speed and space headway data. Section V presents a method for predicting real-time traffic flow parameters using a noise reduction model at low penetration of CVs. Section VI entails an evaluation of the method and the analytical results. Section VII details the real-time application efficacy and concluding remarks are provided in Section VIII.

## II. CONTRIBUTION OF THE PAPER

The contribution of our paper is in the development of a model for the real-time prediction of short-term traffic flow parameters with high accuracy using BSMs at a low CV penetration level. Estimated traffic flow parameters from low penetration of connected vehicles become noisy compared to 100% penetration of CVs and such noise significantly reduce the real-time traffic flow parameters prediction accuracy of a machine-learning model, such as the accuracy of long short-term memory (LSTM) model in terms of predicting speed and space headway data. The real-time prediction model combines a recurrent neural network (i.e., LSTM) with a Kalman filter based noise reduction model. The advantage of this prediction model is that it can accurately predict traffic flow parameters, such as speed and space-headway, with only 5% CV penetration.

## III. RELATED WORK

The related work analyzes existing research on different types of recurrent neural network models for traffic prediction and noise reduction models.

### A. Recurrent Neural Network Models for Traffic flow parameters Prediction

Feed Forward Neural Networks (FFNN), the simplest neural networks, have been used to forecast travel time and traffic flow and subsequent traffic patterns [7], [15], [16]. However, their lack of a memory mechanism prevents capture of temporal and spatial variations in time series problems. Such variations can recall the effect of dynamic nature of traffic patterns, for purposes of mapping future traffic-flow predictions. Further, spatial and temporal patterns and optimal look-back intervals must be determined prior to input into FFNN for the time series prediction. It is required to prepare a large enough input time-series dataset for the FFNN through data preprocessing using statistical methods (e.g., correlation analyses, principal component analysis, and genetic algorithm) to capture spatial and temporal patterns.

RNNs have been used to capture variations over time for such time series problem as the Time-Delay Neural Network (TDNN), the Jordan–Elman Neural Network, and the State-Space Neural Network (SSNN)). The results of the first two models used for traffic speed predictions via 30-s loop-detector speed data from a freeway segment of Interstate 4 in Orlando, Florida indicated their superior performance over the non-linear statistical time series model [17]. The SSNN model was also used for real-time short-term freeway travel time prediction via synthetic and real-world data [18], with one such example the real-time data collected from the freeway and urban scenarios for the Regiolab-Delft Project [19], [20]. However, these NN models cannot capture temporal and spatial relationship for a long-term time series problem due to vanishing gradient and exploding gradient problems.

In addition to these models, there are several variations of RNN models, such as Simple RNN, GRU and LSTM and the difference in the RNN models lies in the transfer function of the repeater block [4]. In the simple RNN model, the transfer function h is merely an activation function. Consequently, the LSTM model was used to address long-term time series problems in terms of traffic flow parameters prediction. For example, Ma et al. used a three-hidden-layer LSTM model for traffic speed prediction utilizing microwave sensor data [4]. The LSTM provided more accurate predictions than traditional RNN models by determining optimal time lags using a trial and error method. The hidden layer of the LSTM model includes a memory block, which contains an input, a forget gate and an output gate, for capturing the non-linear patterns of speed over time. On the other hand, the GRU is a simplified version of the more complex LSTM unit that combines the input and forgets gates into a single update gate. It then merges both the cell and hidden states for faster operation. Thus, Simple RNN and GRU models cannot capture the traffic dynamics accurately.

More recently, Zhao et al. constructed a multi-layers LSTM network for traffic volume prediction. Their model includes an Origin-Destination Correlation (ODC) matrix integrated in the LSTM network [10]. This ODC matrix captures the correlations between the temporal and spatial patterns among different links of a road network, thus improving the performance of the LSTM model by capturing traffic flow evolution over time and space. This study found that the two-dimensional LSTM was more accurate than existing traffic forecast methods for short-term travel speed prediction. In another study, Wang et al. developed a deep neural network using an LSTM for predicting driver behavior [21]. As driver behavior is a time-dependent phenomenon and an LSTM can mimic human memory, they developed an LSTM-based car-following model that can replicate driver behavior using microscopic NGSIM data. This research found that this deep neural network model exhibits significantly higher accuracy than existing car-following models.

However, an LSTM has yet to be used for predicting different traffic flow parameters using BSMs in a connected vehicle environment at a low penetration rate of CVs. As the related work indicates that this type of RNN has the capability of capturing long-term dependency for predicting time series data, this study used an LSTM model for predicting traffic flow parameters in a connected vehicle environment. In this environment, the LSTM model can learn non-linear time-variant traffic behavior from a training data set and predict traffic flow parameters based on the real-time input of traffic flow parameters. However, this learning capability of the LSTM model can be reduced by the noise in the data from a mixed traffic environment (i.e., connected and non-connected vehicles), as one cannot expect 100% CVs in the near future.

*B. Statistical Models for time series prediction*

Aside from RNN models, there are several popular statistical models for short-term time series prediction. In this study, we have used three statistical models as baseline models to compare with RNN based models. The three baseline models are linear regression, Auto-Regressive Integrated Moving Average model (ARIMA) and Holt's Exponential Smoothing (HES) model [22], [23]. These are popular models for time series forecasting. The linear regression model identifies a straight line to fit the time series based on least square principles. The regression model has been implemented using the "Scikit-learn" package in Python [24].

On the other hand, ARIMA is a moving average model that forecasts for future time steps using some number of previous time steps in the dataset. This model has three parameters, the lag order, the degree of differencing and the size of the moving average (MA) window. If one uses seasonal differencing to the time series to make it stationary, we can use the value of the differencing parameter zero. The lag order and MA window size for the ARIMA model can be identified using a grid search method. The values are extracted from the model with the least Akaike Information Criterion (AIC) value [25]. AIC is used to measure the quality of a statistical model compared to other models. The ARIMA model has been implemented using the "Statsmodels" package in Python [26].

Holt's Exponential Smoothing is known as basic/single exponential smoothing technique or EST model. It can be observed that each new prediction depends on all the previous values in the time series with continuously increasing powers of coefficients, Hence it is called exponential smoothing. In this model, a parameter is used as a smoothing factor and it can be anywhere between zero and one. A value closer to one gives more importance to recent observations in time series, whereas a value closer to zero gives more importance to smoothing [27]. The HES model can be implemented using the "Statsmodels" package in Python [26].

*C. Noise Reduction Models*

Noise reduction models have been used extensively to analyze given measurements and to estimate accurate measurements because of the inaccuracies of sensor-collected data. Previously vehicle trajectories data were filtered using the following methods: (i) averaging [28]; (ii) locally weighted regression using the tri-cube weight function [29]; (iii) filtering [14], [30] and (iv) moving average techniques [31]. The noise reduction accuracy of these methods depends on a time window size. Kanagaraj et al. [32] and Rim et al. [33] used locally weighted regression techniques for smoothing erroneous vehicle coordinates and speed data, respectively, both finding that the accuracy of locally weighted regression varies based on the polynomial order. More recently, Punzo et al. [30], [34] used moving average and low pass filtering techniques to correct GPS-based trajectory data, with the latter study analyzing the vehicle trajectory and speed data and evaluating the accuracy in terms of jerk, consistency, and spectral analysis. They found that the low pass filter performs very well in terms of accuracy.

Another widely used data smoothing technique is the Kalman Filter, used to reduce noise from sensor fault in sensor-collected data. This filter, named after Rudolf E. Kalman, who provided the concept for this method [35], estimates the current state based on a sequence of previous noisy observations. There are three types of Kalman filter smoothing, fixed-interval smoothing, fixed-point smoothing, and fixed-lag smoothing in addition to several variations including the standard Kalman filter, the extended Kalman filter and the scented Kalman filter [36]. If the noise in the sensor-collected data is Gaussian, the standard Kalman filter is applicable for the noise reduction. The standard Kalman Filter has been effective in estimating air-vehicle sensor errors. In the past, Ervin et al. used a Kalman filter to smooth the vehicle trajectory data [37].

Using the standard Kalman filter process, Rauch et al. developed an efficient method based on the RTS algorithm, a two-pass algorithm that reduces the computational effort required for Kalman filter smoothing since it requires the standard Kalman filter to be implemented only in the forward direction [38]. The forward pass is the standard Kalman filter while the backward recursion reduces the inherent bias in the Kalman filter estimates. Based on their applicability, the RTS and standard Kalman filter were all used as noise reduction models. In a mixed traffic scenario (connected and non-connected vehicles), data collected from the low penetration of connected vehicles (e.g., 5 % CVs, 10% CVs) in addition to the temporal variation of traffic make the data noisy.

IV. TRAFFIC FLOW PARAMETERS PREDICTION WITH NOISY CONNECTED VEHICLE DATA

In this section, we demonstrate the performance of a state-of-the-art prediction model, i.e., LSTM to predict short-term traffic flow parameters from limited CV data at a low penetration rate of CVs.

*A. Basic Safety Messages (BSMs) Obtained from the Enhanced NGSIM Dataset*

The NGSIM data was collected from each vehicle of a 500m (1650 ft) roadway section on Interstate 80 segment in Emeryville (San Francisco), California (see Figure 1) [14]. The original NGSIM dataset, collected through video cameras, represents 45 minutes of the peak afternoon period, specifically 4:00 PM to 4:15 PM, and 5:00 PM to 5:30 PM. Video image processing was used to generate vehicle trajectory data. However, since the original NGSIM data contain inconsistencies and noise, Montanino and Punzo et al. improved the data set using a multistep procedure to reconstruct the original I 80 dataset (4.00 PM to 4.15 PM) for each vehicle trajectory, and subsequently conducted an extensive exploratory study to determine the accuracy of NGSIM trajectory data [14]. They reconstructed the original data measurements while preserving the actual in-motion vehicle dynamics (i.e., shifting gears, vehicle stoppages), the vehicle trajectory consistency (i.e. acceleration/deceleration, speed, and space headway) and platoon consistency (i.e. the actual space headway between lead and follower vehicles in the traffic stream) [14]. The result is the Enhanced NGSIM dataset.

As this enhanced dataset was collected with a frequency of one-tenth of a second, it represents a sample of the BSMs (i.e., Vehicle ID, Timestamp, Lane ID, Location, Acceleration/Deceleration, Vehicle Length, Vehicle Class ID, Follower Vehicle ID, and Immediate Preceding Vehicle ID) generated from a connected vehicle environment. The study reported here used trajectory data from 3,335 vehicles with a frequency of 10 Hz from the Enhanced NGSIM I80-1 dataset. More specifically, the following subsection section explored the noise and outliers in the speed and space headway data from 5% to 90% penetration of CVs (5%, 10%, 20%, 30%, 40%, 50%, 60%, 70%, 80%, and 90%).

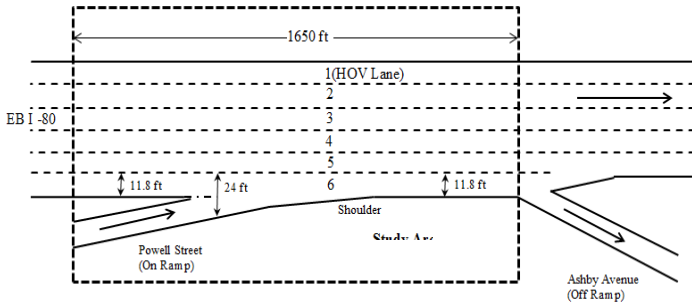

Fig. 1. Study area schematic related to NGSIM data (adapted from [14]).

### B. Identification of Noise in the Speed and Space Headway Data

Let's assume that CV penetration is $q\%$. This means that at each time step, we randomly sample $q\%$ vehicle IDs from the available vehicle IDs at time step $t$ and then take the average (speed and headway) of those $q\%$ vehicles for which the data is available at time step $t$. This method of aggregation ensures that we avoid the issue of discontinuity in the time series. The BSMs in this study include the timestamp, location coordinates, speed, acceleration/deceleration, relative speed, lane number, leader vehicle number and follower vehicle number. To prepare the data as a time series problem, we used the Frame ID sequence from the NGSIM data as each video frame was created every one-tenth of a second. The space headway of a vehicle was then calculated using the location coordinate of each vehicle. The average speed and space headway time-series data extracted from BSMs at different penetration rates of CVs to identify noise and the type of noise distribution. This low CV penetration prevents the acquisition of accurate average speed and space headway data of all vehicles in a roadway segment. Further, data from limited CVs hinder an accurate estimation of current average speed or space headway, thus yielding a large number of outliers in the average traffic flow parameters. The box plots of the average speed and space headway data are in Figures 2(a) and 2(b) respectively, with varying penetrations of CVs. The number of outliers increases in both average speed and space-headway data with a decreasing in CV penetration rates, indicating a spatial variation in both speed and headway data. The variation of speed and headway over time for both penetration levels is shown in Figure 2. In Figures 2(a) and 2(b), outliers occur because the speeds and headways are not uniform across the study area as shown in Figure 1, which indicates the spatial variation of speed and space headway.

To observe the change in traffic flow parameters over time, we compared the average speed and space headway profile between 100% penetration of CVs and 10% penetration of CVs in Figures 3(a) and 3(b), respectively. As indicated, the average speed and space-headway change drastically over time with a 10% CV penetration rate compared to a 100% CV penetration. Both the speed and space headway estimation from the low penetration of CVs include error compared to 100% penetration of CVs and we define this error as a noise. This noise leads to an inaccurate prediction of traffic flow parameters in any models including machine learning models. In addition, a massive amount of data is required to train a deep learning model to develop a prediction model with this noisy traffic flow parameters used to capture the variation of the traffic behavior and to achieve expected accuracy. However, it is possible to achieve an expected prediction accuracy with a low penetration rate of CVs if we can reduce the noise in the traffic flow parameters.

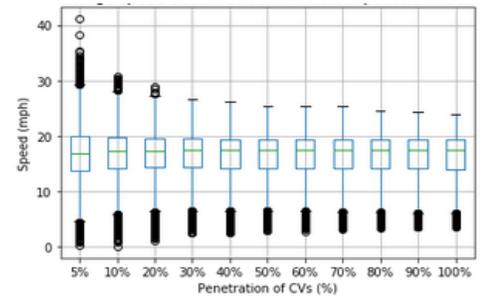

(a) Average speed distribution

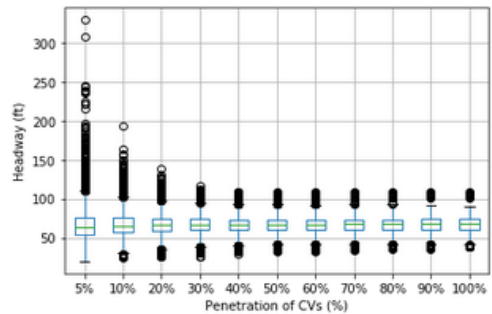

(b) Average space headway distribution

Fig. 2. Average speed and space headway distribution with varying penetration of CVs.

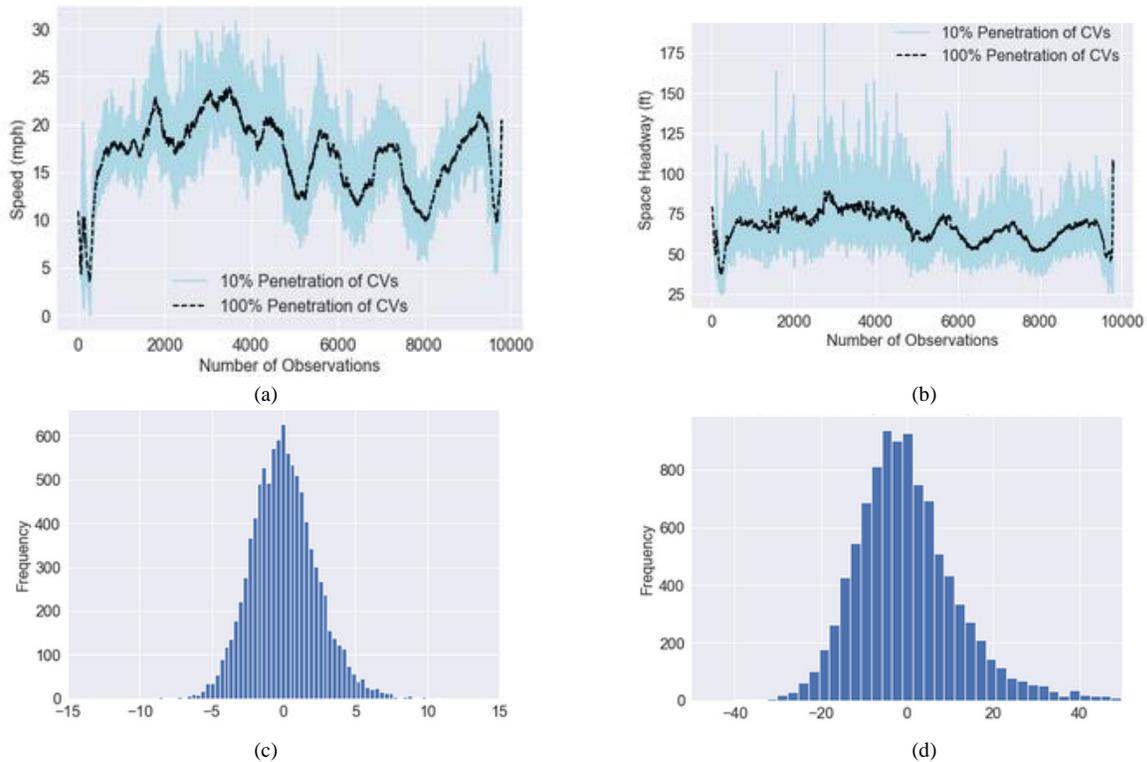

Fig. 3. Noise in speed and space headway distribution for 10% CV penetration: (a) Comparison between 10% and 100 % CV penetration for speed data; (b) Comparison between 10% and 100 % CV penetration for space headway data; (c) Noise distribution histogram in speed data for 10% CV penetration; and (d) Noise distribution histogram in space headway data for 10% CV penetration.

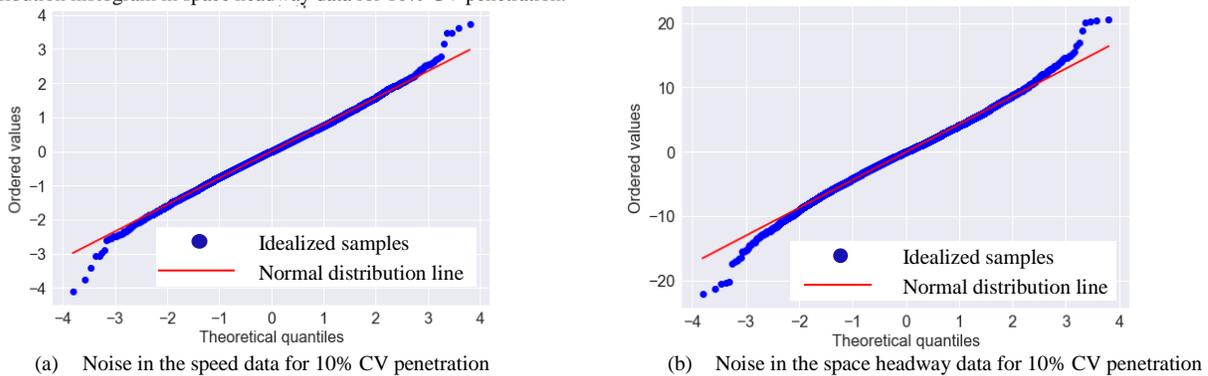

(a) Noise in the speed data for 10% CV penetration     (b) Noise in the space headway data for 10% CV penetration

Fig. 4. An example of a Q-Q plot generated for noise in the average speed and space headway data.

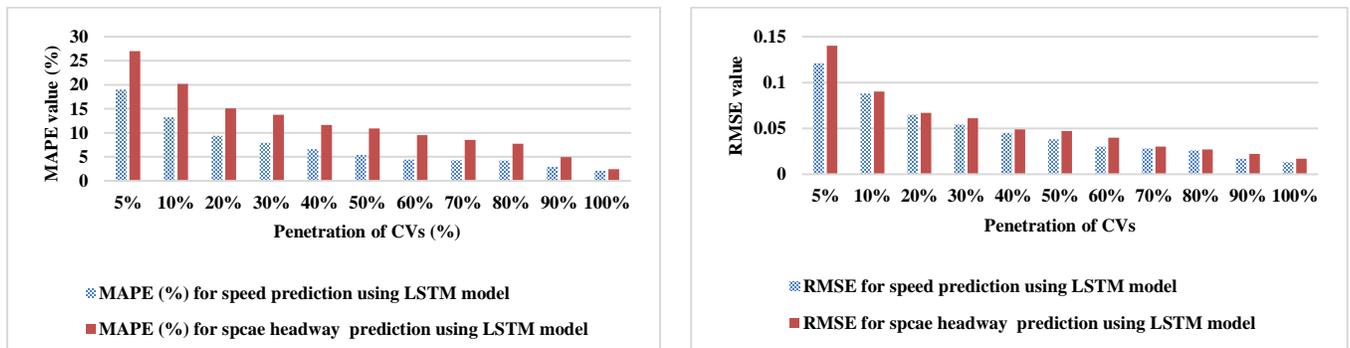

(a) MAPE value for real-time speed and space headway prediction accuracy of the LSTM model     (b) RMSE value for real-time speed and space headway prediction accuracy of the LSTM model

Fig. 5. Comparison between MAPE and RMSE values related to real-time speed and space headway prediction accuracy of LSTM model for different CV penetration rates.

We identify the type of the distribution to reduce noise (i.e., inaccuracy) in the estimated current traffic flow parameters, as the selection of the noise reduction model depends on the noise distribution. For example, using the standard Kalman Filter and RTS noise reduction model, it is possible to reduce the noise if the noise distribution in the data is Gaussian. Figures 3(c) and 3(d) present the histogram of the noise distributions in the speed and space headway data. To analyze the type of noise distributions, the noise for each observation was calculated by subtracting the speed or space headway data for the 10% penetration from the 100% penetration levels of CVs. This process was followed for all ten CV penetration rates studied here. We found that the noise distributions of all penetration levels of CVs followed a Gaussian distribution (i.e., normal distribution). In addition, the analysis of the distributions indicates that the noise distributions for all penetration rates followed a similar distribution.

To confirm the normality (Gaussian distribution) of the noise, a quantile-quantile plot (Q-Q plot) was used for further data analysis. In this study, Q-Q plot generated idealized samples based on the Gaussian or normal distribution from the given noise. These idealized samples were divided into groups called quantiles. Each data point in the sample was paired with a similar member from the idealized distribution at the same cumulative distribution, and the resulting points were plotted as a scatter plot with the idealized value on the x-axis and the data sample on the y-axis. The resulting plot indicated that the idealized samples followed the normal distribution lines, confirming that the noise distributions for all penetration rates followed a normal distribution. Figure 4 presents an example of one of the Q-Q plots generated for noise in the average speed and space headway data. Before developing a prediction model using noisy traffic flow parameters, we evaluate the prediction performance of a widely used traffic flow parameters prediction model (i.e., LSTM) at a low penetration rate of CVs. For this purpose, we used Enhanced NGSIM dataset to evaluate the performance of the LSTM model. The following sub-sections provide detail related to training and testing and performance evaluation of the LSTM model.

### C. Training and Testing of the LSTM model

After processing the BSMs from the enhanced NGSIM dataset for different penetration rate for CVs, the time series of noisy speed and space headway data were used as input for a supervised learning problem. Specifically, we used the observation of speed or space headway at the current time step as an input to predict the traffic observation at the next time step. To develop and evaluate the LSTM model, the dataset was divided into a training dataset containing 7000 samples and a testing dataset of 2800 samples for a total of 9800 samples. We have used optimal values of LSTM hyperparameters (number of epochs = 400; number of neurons = 100; batch size = 50; dropout rate = 0.2; and learning rate = 0.001) for training and testing of the model. However, we found that it is impossible to fit a model with the noisy and limited amount of data. It should be noted that for training and testing, the data were normalized between zero and one prior to use as an LSTM model input. After predicting the traffic flow parameters, we calculated the Root Mean Square Error (RMSE) based on the difference between the scaled value of predicted traffic flow parameters and the scaled value of actual traffic flow parameters (ground truth).

### D. Impact of Noisy Data in the Traffic flow parameters Prediction Accuracy

The noise in the traffic flow parameters - because of low penetration of CVs - has a significant impact on the performance of the traffic flow parameters prediction, i.e., speed and space headway. Figure 5 shows that accuracy is decreasing with decreasing the penetration of CVs in terms of Mean Absolute Percentage Error (MAPE) and Root Mean Square Error (RMSE). We then calculated all RMSE and MAPE values based on normalized ground truth and the predicted value of speed using the optimal hyperparameters for each of the prediction models. Table I summarizes the statistical significance test to identify significant differences between the predicted data and the actual data. For this purpose, the t-test was conducted at a 95% confidence interval, the results of which indicated significant differences of the predicted speed using only LSTM is with the actual value for CV penetration ranging from 5% to 50%, and that of the space headway from the actual value for the 5% to 60% penetration range. This indicates that the LSTM model cannot predict traffic flow parameters accurately because of noise in the data at low penetration of CVs. Thus, the focus of our paper is to develop a prediction model that can predict traffic flow data at low penetration of CVs.

TABLE I
SUMMARY OF STATISTICAL SIGNIFICANCE TEST

| Traffic flow parameters | Penetration of Connected Vehicles | | | | | | | | | | |
|---|---|---|---|---|---|---|---|---|---|---|---|
| | 5% | 10% | 20% | 30% | 40% | 50% | 60% | 70% | 80% | 90% | 100% |
| | Baseline model (LSTM without noise reduction filter) | | | | | | | | | | |
| Speed | × | × | × | × | × | × | √ | √ | √ | √ | √ |
| Space headway | × | × | × | × | × | × | × | √ | √ | √ | √ |

**Note:** × = the actual and predicted values significantly different with 95% confidence interval; √ = the actual and predicted values are not significantly different at 95 % confidence interval.

## V. REAL-TIME TRAFFIC FLOW PARAMETERS PREDICTION MODEL WITH NOISE CORRECTION

To reduce the noise from the limited data at low CV penetration rates, the general traffic-data prediction model framework developed in this study that uses LSTM combined with the noise reduction model is shown in Figure 6. Noise reduction models filter noise from the average traffic flow parameters in real-time. The traffic flow parameters sequence for the noise reduction model input is denoted as $\boldsymbol{x} = (x_1, x_2, x_3, \ldots, x_{t-2}, x_{t-1}, x_t)$, and the noise reduction model output sequence is denoted as $\boldsymbol{h}^f = (h_1^f, h_2^f, h_3^f, \ldots, h_{t-2}^f, h_{t-1}^f, h_t^f)$, where $t$ is the total number of time-steps. In the context of traffic flow parameters prediction, $x_t$ is traffic flow parameters at time-step t, and $h_t^f$ is the de-noised traffic flow parameters at time-step t. For real-time traffic flow parameters predictions, de-noised traffic flow parameters, $h_t^f$ is the input for the LSTM model at time-step t and $h_{t+1}^p$ is the predicted traffic flow parameters at time-step $(t+1)$. The enhanced NGSIM data collected at every tenth of a second for each vehicle was used as a part of the BSMs in a connected vehicle environment. The following subsections describe in detail the LSTM and noise reduction model.

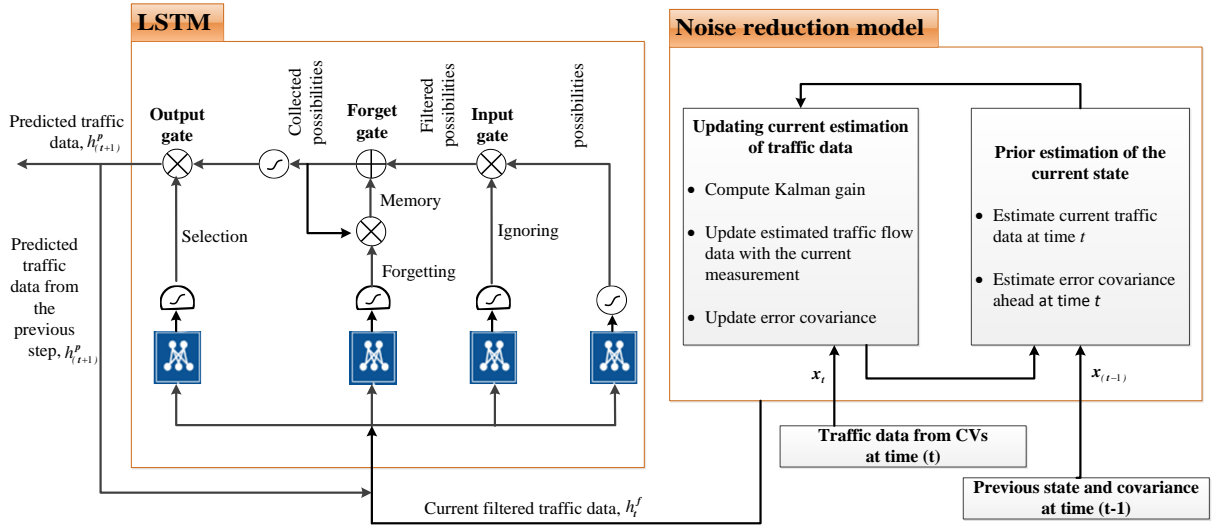

Fig. 6. Real-time prediction model for traffic flow parameters at low penetration of CVs.

*A. Noise Reduction Model*

We used the Standard Kalman filter and the Kalman filter based RTS noise reduction models to filter the noise in the average speed data and for the space-headway filtering.

*1) Standard Kalman Filter*

Based on the Kalman filter [35], the state of speed ($x_t$) at time-step $t$ evolves from the state ($x_{t-1}$) at time-step $(t-1)$.

$$x_t = A_t x_{t-1} + B_t u_t + w_t \quad (1)$$

where $A_t$ is the state transition matrix that transforms state $x_{t-1}$ to state $x_t$; $B_t$ is the control-input matrix for measuring the correction for external influences based on control vector $u_t$; and $w_t$ is the process noise, which is assumed to be drawn from a zero mean multivariate normal distribution $N$ with covariance $Q_t$: $w_t \sim N(0, Q_t)$.

At time-step $t$, $z_t$ is a measurement value calculated based on the linear combination of the newly estimated speed $x_t$ and the measurement noise $v_t$.

$$z_t = H_t x_t + v_t \quad (2)$$

Where $H_t$ is the measurement matrix that transforms the new estimated state of speed to a measured state and $v_t$ is the measurement noise, which is assumed to be zero-mean Gaussian white noise with covariance $R_t$: $v_t \sim N(0, R_t)$.

The Kalman filter algorithm consists of two stages for reducing the noise of the speed data: i) prior estimation of the new state and ii) measurement update. Using the following equations, new speed are estimated at time-step $t$. Here, $P$ is a prior or posterior error covariance matrix, which measures the estimated accuracy of the state estimate.

$$\hat{x}_{t|t-1}^{prior} = A_t \hat{x}_{t-1|t-1} + B_t u_t \quad (3)$$

$$P_{t|t-1}^{prior} = A_t P_{t-1|t-1} A_t^T + Q_t \quad (4)$$

Next, the prior estimation of speed $\hat{x}_{t|t-1}^{prior}$ and covariance $P_{t|t-1}^{prior}$ are required for updating the measurement at time-step $t$. The current speed $\hat{x}_t$ is thus estimated at time-step $t$:

$$\hat{x}_{t|t} = \hat{x}_{t|t-1}^{prior} + K_t \left( z_t - H_t \hat{x}_{t|t-1}^{prior} \right) \quad (5)$$

where $K_t$ is the Kalman gain. Covariance $P_t$ is calculated for updating the value of $x$ at time-step $(t-1)$ as follows:

$$P_t = \left( I - H_t K_t \right) P_{t|t-1}^{prior} \quad (6)$$

Using the prior covariance $P_t^{prior}$, we calculate the Kalman gain as follows:

$$K_t = P_{t|t-1}^{prior} H_t^T \left( H_t P_{t|t-1}^{prior} H_t^T + R_t \right)^{-1} \quad (7)$$

*2) Kalman Filter Based Rauch–Tung–Striebel (RTS) Model*

The Rauch–Tung–Striebel (RTS) smoother uses the same forward pass as the standard Kalman filter algorithm [38]. The resulting prior and posterior speed estimates $\hat{x}_{t|t-1}$ and $\hat{x}_{t|t}$, and covariances $P_{t|t-1}$ and $P_{t|t}$ from the forward pass are used in the backward pass, which computes the smoothed speed estimates $\hat{x}_{t|n}$ and covariance $P_{t|n}$ (where $t<n$). The backward steps are next completed using the following recursive equations:

$$\hat{x}_{t|n} = \hat{x}_{t|t} + C_t \left( \hat{x}_{t+1|n} - \hat{x}_{t+1|t} \right) \quad (8)$$

$$P_{t|n} = P_{t|t} + C_t \left( P_{t+1|n} - P_{t+1|t} \right) C_t^T \quad (9)$$

where, $C_t = P_{t|t} A_{t+1}^T P_{t+1|t}^{-1}$

where $x_{t|t}$ is the posterior speed estimate of time-step $t$; $A_{t+1}^T$ is the transpose of the state transition matrix at $t+1$; $x_{t+1|t}$ is the prior speed estimate of time-step $t+1$; $P_{t|t}$ is the posterior covariance estimate of time-step $t$, and $P_{t+1|t}$ is the prior covariance estimate of time-step $t+1$.

### B. Long Short-Term Memory (LSTM)

De-noised data from a noise reduction model will be used as an input in the LSTM model. The LSTM model used in this research consists of (i) an input layer, (ii) a recurrent hidden layer, and (iii) an output layer [13]. The input sequence for the input layer is denoted as $x = (x_1, x_2, x_3, ..., x_t)$, and the output sequence for the output layer is $h = (h_1, h_2, h_3, ..., h_t)$, where $t$ is the total number of time-step. In the context of speed and space headway prediction, $x$ is the current speed or space headway data, and $h$ is the predicted speed. Of these layers, the primary layer is the recurrent hidden layer, which consists of a memory block, which solves the vanishing gradient (i.e., a change in the current speed or space headway causes very small change of the predicted speed or space headway) or exploding gradient (i.e., a change in the current speed or space headway causes very big change of the predicted speed or space headway) problems of traditional RNNs.

The memory block consists of a forget gate, an input gate, and an output gate (as shown in Figure 6), all three of which control what information needs to be removed or added from the previous cell state to the new cell state. The input gate controls the activations of input into the memory block. The input gate $i_t$ decides the requisite values requiring an update using a sigmoid activation function:

$$i_t = sigmoid(w_i x_t + u_i h_{t-1} + b_i) \quad (10)$$

where $w$ and $u$ are the parameter matrices, and $b$ is the bias. The forget gate determines the information that must be forgotten from the previous cell state. Using a sigmoid layer, the forget gate layer $f_t$, which is represented by the following equation, determines the information to forget.

$$f_t = sigmoid(w_f x_t + u_f h_{t-1} + b_f) \quad (11)$$

We then use the input gate and forget gate information to update the previous cell state, $c_{t-1}$, to new cell state $c_t$. To obtain the new cell state, we multiply the previous cell state $c_{t-1}$ is multiplied by $f_t$ to forget unnecessary information from the previous state. We can thus add new candidate values $i_t \odot tanh(w_c x_t + u_c h_{t-1} + b_c)$ to define that which is needed to update each state value:

$$c_t = f_t \odot c_{t-1} + i_t \odot tanh(w_c x_t + u_c h_{t-1} + b_c) \quad (12)$$

The output gate controls the activations of output into the memory block. At the output gate, a sigmoid layer decides what parts of the cell state to output, $o_t$:

$$o_t = sigmoid(w_o x_t + u_o h_{t-1} + b_o) \quad (13)$$

We then put cell state $c_t$ through $tanh$ (to push the values to between −1 and 1) activation functions and multiplied by the output of the sigmoid gate output $o_t$ to predict speed or space headway $h$:

$$h_t = o_t \odot tanh(c_t) \quad (14)$$

However, the prediction accuracy of the LSTM model depends on the determination of the optimal hyperparameter that includes the number of neurons, the number of epochs, the batch size, the dropout rate, and the learning rate.

### C. Optimal Hyperparameter Determination of the LSTM

For the time series problem, traditional hyperparameter selection methods such as the grid search method and the random search method [39] are inapplicable for determining the optimal hyperparameter set. We thus used a trial-and-error procedure and the Root Mean Square Error (RMSE) metric to determine the optimal LSTM hyperparameter set. RMSE measures the square root of the average of the squared errors, which quantifies the difference between the predicted values and the actual values. The mathematical formulation of RMSE is as follows:

$$RMSE = \frac{1}{N}\sqrt{\sum_{i=1}^{N}(y_i - \hat{y}_i)^2} \quad (15)$$

where $N$ represents the total sample size, $y_i$ is the actual value of traffic flow parameters (speed or space headway) and $\hat{y}_i$ is the predicted value of traffic flow parameters (predicted speed or predicted space headway).

Next, we used a box and whisker plot to identify the optimal parameter set for a specific hyperparameter and compare the distribution of the RMSE scores for the various hyperparameter values. Figure 7 shows the box and whisker plot for the number of the neurons selection process for the LSTM model for 5% penetration of CVs. We created and plotted thirty samples for each value of a hyperparameter in the box and whisker plot to select the optimal hyperparameter set. The plot shows the median (green line), 25[th] and 75[th] percentiles of the data. This comparison also indicates that the optimal number of neurons is 100. However, the plot also shows that we could achieve better instantaneous performance at the cost of worse average performance. A similar procedure was followed for the selection of other hyperparameters (i.e., number of epoch, batch size, dropout rate, learning rate) for different penetrations of CVs. The optimal hyperparameter values for LSTM model are number of epochs = 400; number of neurons = 100; batch size = 50; dropout rate = 0.2; and learning rate = 0.001.

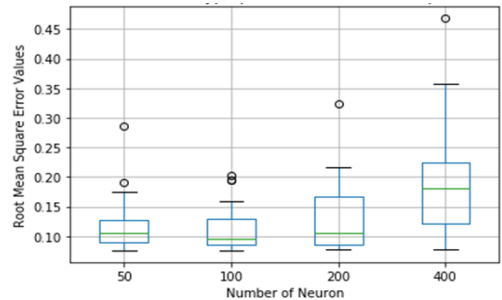

Fig. 7. Box and whisker plot for tuning the number of neuron for 5% penetration of CVs.

### D. Training and Testing of the Combined Model

To train the LSTM model, we used a stochastic gradient descent algorithm with adaptive learning rate tricks [40] and an optimal hyperparameter estimated as described in the previous section. The advanced gradient descent algorithm, ADAM, is an extension of the existing stochastic gradient descent algorithm [40], which was used because of its applicability in natural language processing applications. After training the model using optimal hyperparameters and ADAM, we then determined the goodness of fit of the LSTM model to that of the testing datasets by determining the overfitting and underfitting problems to ensure the predictive capability of the model. An identical error occurring in both the training and testing stages of the prediction model represents the model goodness-of-fit. A reduced level of error on the testing rather than training dataset indicates an underfit model; a reduced level of error on the training data set, indicates an overfit model that is characterized by continuous improvement, followed by a plateau and degradation of the testing set. If the loss (or error) on training and testing datasets decreases and stabilizes around the same point, then the model exhibits a good fit with the training and testing data. To assess the goodness-of-fit of the model using the training and testing datasets, the Mean Absolute Error (MAE), defined as the average of the absolute error, is used as a measurement of the loss (i.e., error) and as a metric to evaluate the performance. The mathematical formulation of MAE is given below.

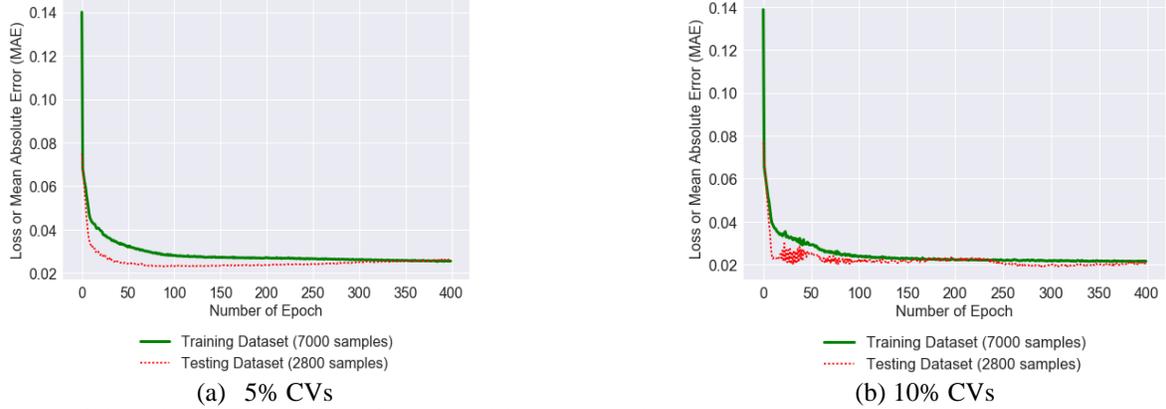

(a) 5% CVs  (b) 10% CVs

Fig. 8. Comparison of Mean Absolute Error (MAE) profiles using training and testing datasets with the optimal parameter set.

$$MAE = \frac{1}{N}\sum_{i=1}^{N}|y_i - \hat{y}_i| \quad (16)$$

where $N$ represents the total sample size; $y_i$ is the actual value of traffic flow parameters (speed or space headway) and $\hat{y}_i$ is the predicted value of traffic flow parameters (predicted speed or predicted space. We plotted MAE profiles for training and testing dataset of different penetration of CVs to evaluate the goodness-of-fit of the LSTM model. The MAE profiles using optimal hyperparameters is show in Figure 8 with both the training and testing datasets from 5% to 30% penetration of CVs. A comparison of the MAE values of these two datasets indicates a good fit of each model with the optimal hyperparameters.

## VI. EVALUATION RESULTS AND DISCUSSIONS

In the evaluation of the traffic flow parameters prediction model, we used a combined model that reduces noise in the data in real-time to predict the speed and space headway. To demonstrate the efficacy of real-time data de-noising strategy at low penetration of CVs, we compared the average speed and space headway profiles with 100% CV penetration rates to de-noised speed and space headway profiles with different CV penetration rates. Figure 9 presents a comparison between average speed profile and space headway profile with 100% CV penetration rate and filtered speed and space headway profiles with 10% CV penetration rate using standard Kalman and RTS filters. We observed an improved level of performance of the RTS filter than the standard Kalman filter to de-noise data at a 10% CV penetration rate. The filtered average speed and space headway profiles with 10% penetration of CVs follow the 100% penetration CV profile very closely. We then quantitatively evaluated filter performance through the prediction accuracy of the average speed and space headway using the LSTM model at low penetration of CVs. We followed the same procedure to de-noise the average speed and space headway data for all CV penetration rates.

We next compared the performance of the traffic flow parameters prediction accuracy between LSTM/RTS, LSTM/Standard Kalman Filter, LSTM/moving average and the baseline model (i.e., only LSTM model using noisy data). Both the Mean Absolute Percentage Error (MAPE) and RMSE serve as the performance evaluation metrics. We omit the mathematical RMSE formulations as they are previously provided and provide here the mathematical MAPE formulation:

$$MAPE = \frac{100\%}{N}\sum_{i=1}^{N}\left|\frac{y_i - \hat{y}_i}{y_i}\right| \quad (17)$$

where $N$ represents the total sample size; $y_i$ the actual value of traffic flow parameters (speed or space headway) and $\hat{y}_i$ the predicted value of traffic flow parameters (either predicted speed or space headway).

In Figure 10, we quantitatively evaluated the performance of the predicted speed and space headway in terms of RMSE. We calculated all RMSE values based on normalized ground truth and the predicted value of speed using the optimal hyperparameters for each prediction models as detailed in Section V.D. The speed data normalization scale range is set from zero (0) to one (1). As shown in Figure 10, a comparison of the RMSE values indicates that the LSTM combined with the three noise reduction models performs better than the LSTM alone. More specifically, the LSTM combined with the RTS filter provided lower RMSE values compared to the Moving Average and the standard Kalman filter.

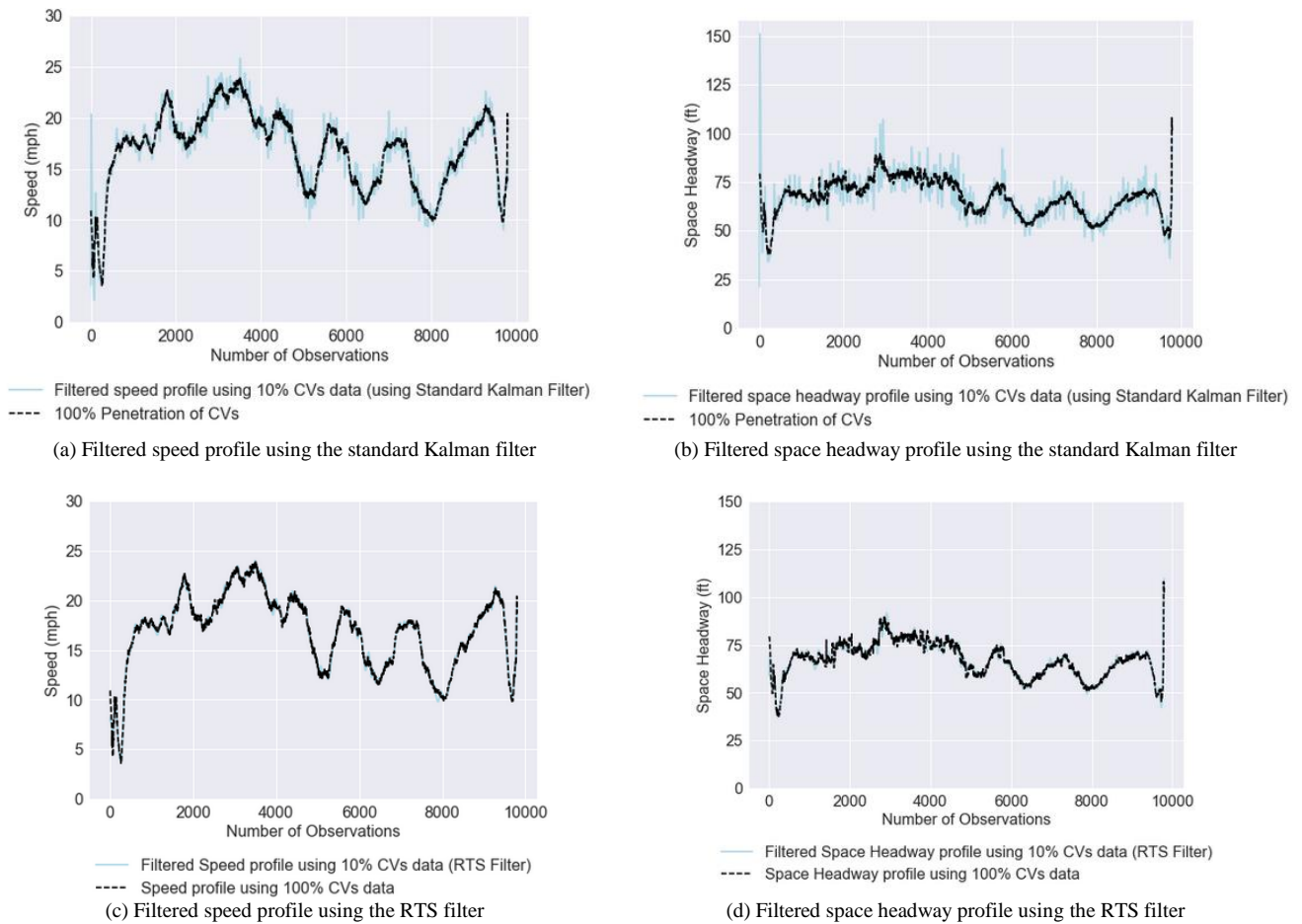

Fig. 9. Comparison between average speed profile (and space headway profile) with a 100% CV penetration rate and filtered speed and space headway profiles with 10% CV penetration rate using standard Kalman Filter and RTS filters.

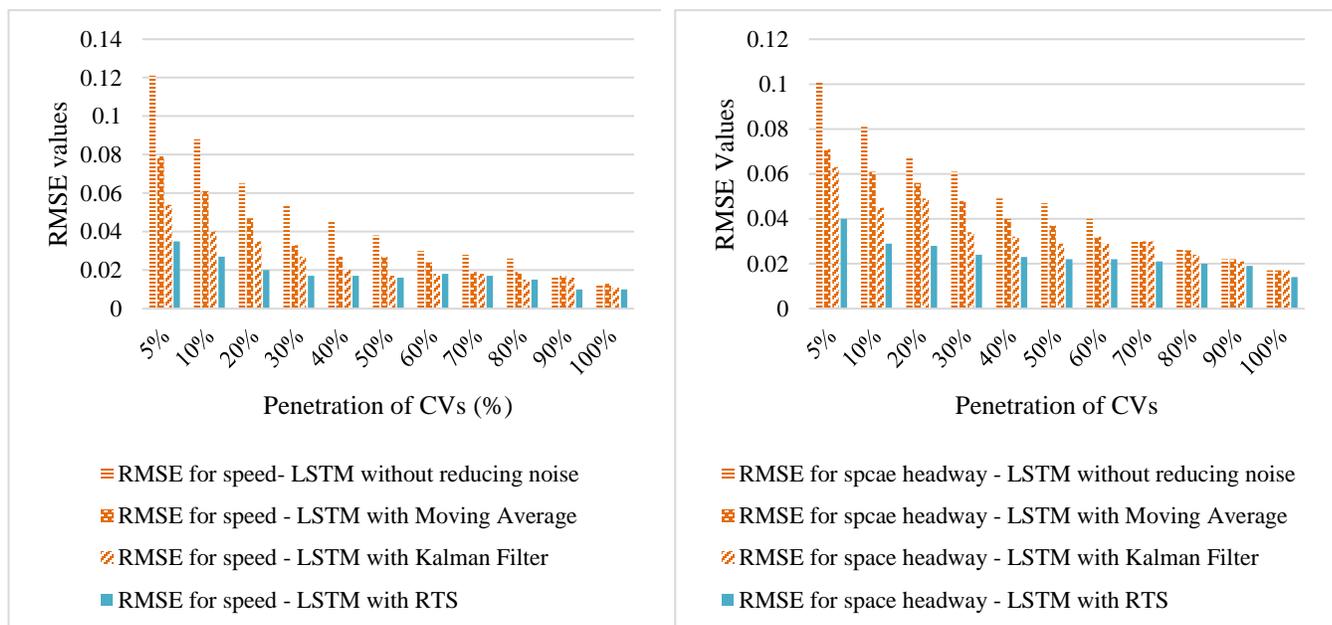

Fig. 10. (a) Comparison of RMSE for speed prediction between the LSTM with the Moving Average, standard Kalman and RTS noise reduction filters, and LSTM without the noise reduction filter.

Fig. 10. (b) Comparison of RMSE for the space headway prediction between the LSTM with the Moving Average, standard Kalman and RTS noise reduction filters, and the LSTM without noise reduction filter.

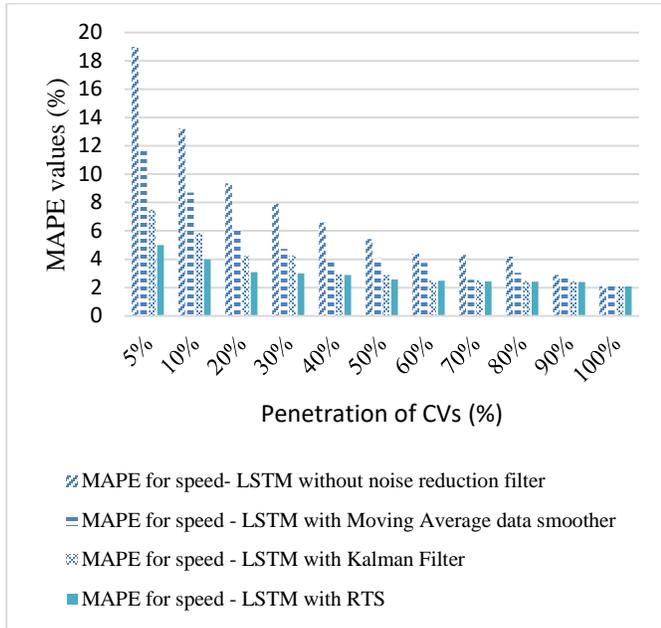

Fig. 11. (a) Comparison of MAPE for speed prediction between the LSTM with the Moving Average, standard Kalman and RTS noise reduction filters, and LSTM without the noise reduction filter.

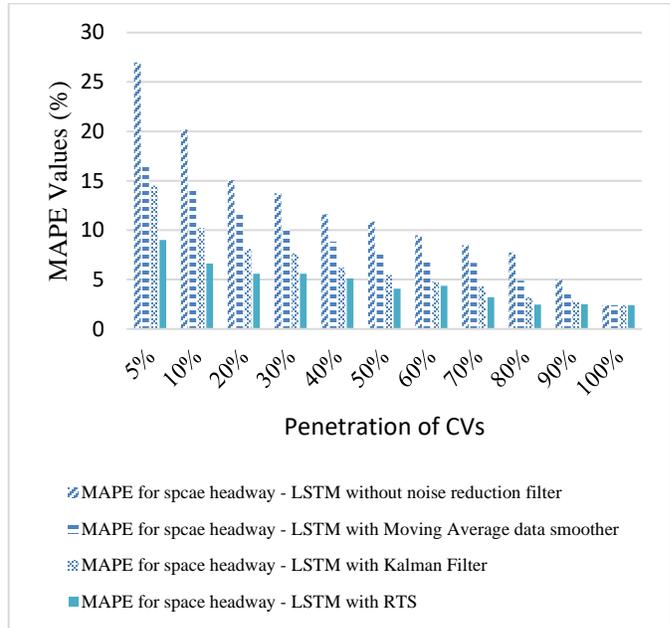

Fig. 11. (b) Comparison of MAPE for the space headway prediction between the LSTM with the Moving Average, standard Kalman and RTS noise reduction filters, and the LSTM without noise reduction filter.

TABLE II
SUMMARY OF STATISTICAL SIGNIFICANCE TEST

| Traffic flow parameters | Penetration of Connected Vehicles | | | | | | | | | | |
|---|---|---|---|---|---|---|---|---|---|---|---|
| | 5% | 10% | 20% | 30% | 40% | 50% | 60% | 70% | 80% | 90% | 100% |
| | Traffic flow parameters prediction using LSTM without reducing noise | | | | | | | | | | |
| Speed | × | × | × | × | × | × | √ | √ | √ | √ | √ |
| Space headway | × | × | × | × | × | × | × | √ | √ | √ | √ |
| | Traffic flow parameters prediction using LSTM with moving average model | | | | | | | | | | |
| Speed | × | × | × | √ | √ | √ | √ | √ | √ | √ | √ |
| Space headway | × | × | × | × | √ | √ | √ | √ | √ | √ | √ |
| | Traffic flow parameters prediction using LSTM with Standard Kalman Filter Model | | | | | | | | | | |
| Speed | × | × | √ | √ | √ | √ | √ | √ | √ | √ | √ |
| Space headway | × | × | × | √ | √ | √ | √ | √ | √ | √ | √ |
| | Traffic flow parameters prediction using LSTM with RTS model | | | | | | | | | | |
| Speed | √ | √ | √ | √ | √ | √ | √ | √ | √ | √ | √ |
| Space headway | √ | √ | √ | √ | √ | √ | √ | √ | √ | √ | √ |

**Note:** × = the actual and predicted values significantly different with 95% confidence interval; √ = the actual and predicted values are not significantly different at 95 % confidence interval.

In Figure 10 (b), we compare the RMSE values for the space headway prediction with the three noise reduction models. As with the speed prediction, this LSTM combination with all three noise-reduction models performs better than the baseline model alone. Furthermore, this LSTM/RTS filter combination yielded the lowest RMSE values similar to the speed prediction.

Unlike the LSTM alone, this combined LSTM/RTS model reduced MAPE from 19% to 5% for the speed prediction at a 5% CV penetration rate as shown in Figure 11(a). Considering all CV penetration rates, the MAPE values using the combined LSTM/RTS model ranges from 1% to 5% for speed predictions. On the other hand, compared to the baseline model, this LSTM/RTS model combination reduced the MAPE from 27% (using LSTM only) to 9% for a space headway prediction with a 5% CV penetration rate. As shown in Figure 11(b), the MAPE values for LSTM/RTS model combination ranges from 2% to 9% for space headway prediction with different CV penetration rates.

IEEE TRANSACTIONS ON INTELLIGENT TRANSPORTATION SYSTEMS 12_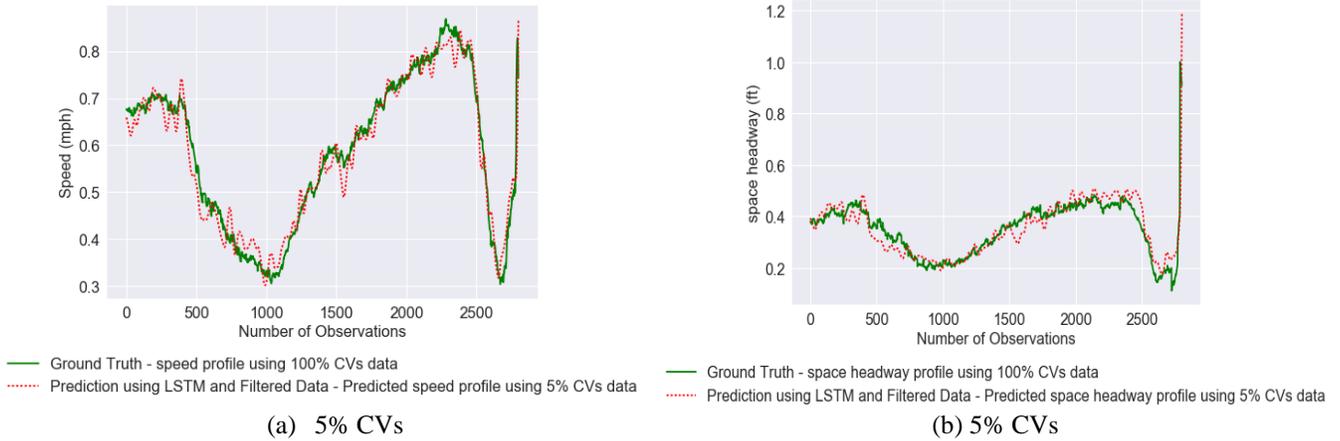

(a) 5% CVs             (b) 5% CVs

Fig. 12. Comparison of the ground truth and predicted speed data using the LSTM combined with RTS for 5% CV penetration rates.

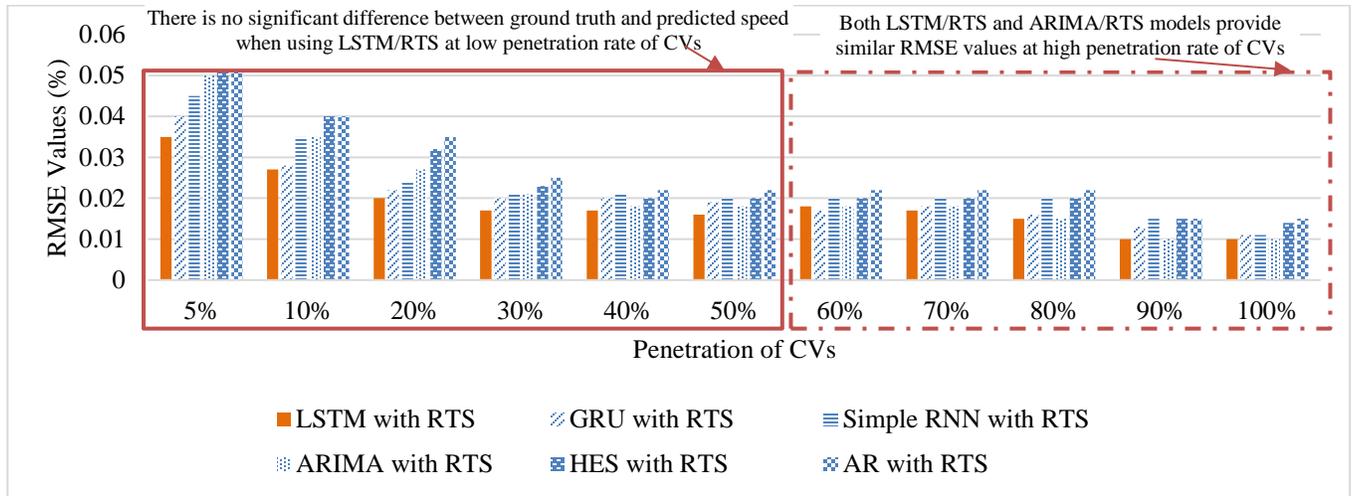

Fig. 13. Comparison of RMSE values for speed predictions between different prediction models combined with RTS.

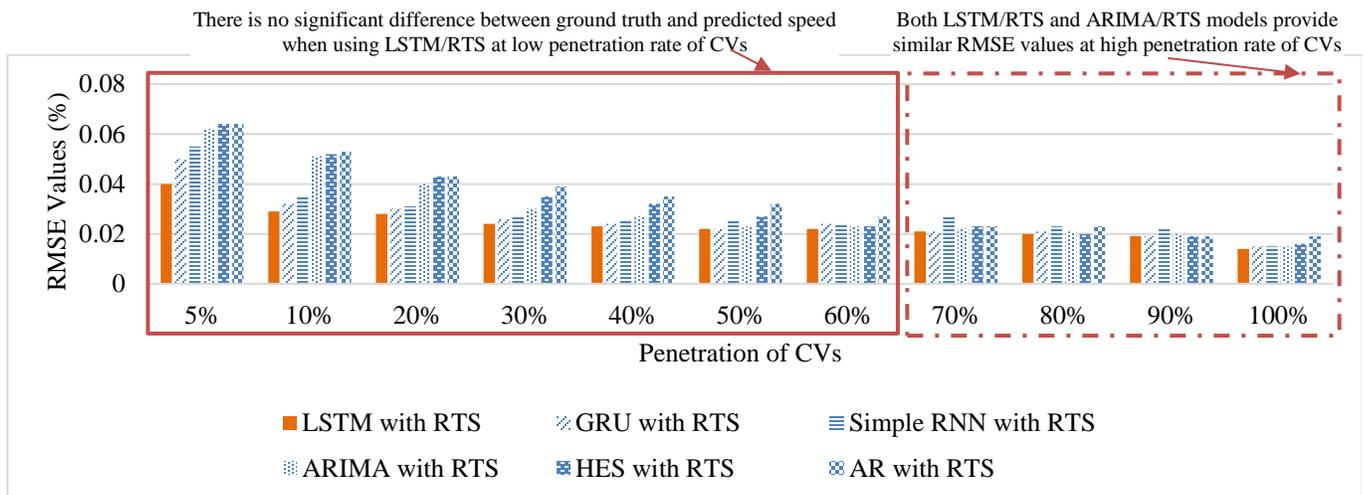

Fig. 14. Comparison of RMSE values for space headway prediction between different prediction models combined with RTS.

TABLE III
SUMMARY OF RMSE, MAE AND MAPE VALUES USING LSTM COMBINED WITH RTS

| Traffic flow parameters | Measure of Effectiveness | Penetration of Connected Vehicles | | | | | | | | | | |
|---|---|---|---|---|---|---|---|---|---|---|---|---|
| | | 5% | 10% | 20% | 30% | 40% | 50% | 60% | 70% | 80% | 90% | 100% |
| | | LSTM with RTS | | | | | | | | | | |
| Speed | RMSE | 0.035 | 0.027 | 0.025 | 0.017 | 0.018 | 0.016 | 0.018 | 0.017 | 0.015 | 0.01 | 0.01 |
| | MAPE (%) | 4.99 | 4 | 3.1 | 3 | 2.9 | 2.57 | 2.49 | 2.45 | 2.42 | 2.4 | 2.09 |
| Space headway | RMSE | 0.04 | 0.029 | 0.028 | 0.024 | 0.028 | 0.022 | 0.024 | 0.027 | 0.02 | 0.019 | 0.014 |
| | MAPE (%) | 9.02 | 6.62 | 5.6 | 5.6 | 5.1 | 4.07 | 4.36 | 3.2 | 2.48 | 2.49 | 2.4 |

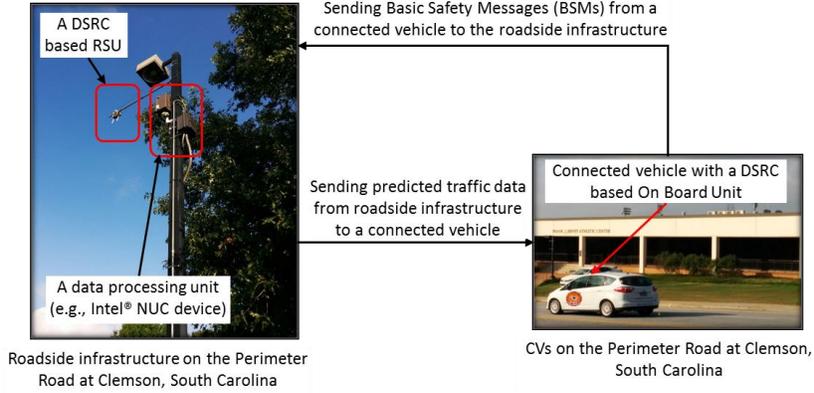

Fig. 15. Field experiments of communication latency between RSU and connected vehicles.

TABLE IV
SUMMARY OF TWO-WAY COMMUNICATION LATENCY AND COMPUTATION TIME FOR THE REAL-TIME TRAFFIC FLOW PARAMETERS PREDICTION APPLICATION

| Two-way communication latency between CVs and RSU using DSRC | Computation time for running the prediction model | Total latency including two-way communication and computation time | Minimum Latency Requirements for Mobility and Environmental Application [44] |
|---|---|---|---|
| 22 ms (Maximum Latency) | 70 ms | 92 ms | ≤ 1000 ms |
| 9 ms (Average Latency) | 70 ms | 79 ms | |
| 5 ms (Minimum Latency) | 70 ms | 75 ms | |

Table II summarizes the statistical significance test to identify significant differences between predicted traffic flow parameters using with different models, i.e., LSTM without reducing noise, LSTM with moving average model, LSTM with Standard Kalman Filter Model, LSTM with RTS model, and the actual data (i.e., ground truth data). For this purpose, the t-test was conducted at a 95% confidence interval, the results of which indicated significant differences of the predicted speed using only LSTM is with the actual value for CV penetration ranging from 5% to 50%, and that of the space headway from the actual value for the 5% to 60% penetration range. No significant difference was observed between the predicted speed and space headway values using LSTM combined with RTS from the actual value at 5% CV penetration.

In Figure 12, we compare the ground truth data (actual speed/space headway using 100% connected vehicles) and the predicted speed and space headway profile using the LSTM/RTS combinations with 5% CV penetration rate as our example. We have also compared the performance of the LSTM/RTS model with popular variation of RNN models (i.e., GRU and Simple RNN) and statistical models (i.e., ARIMA, HES and AR models) to prove the efficacy of LSTM/RTS model. Figures 13 and 14 present a comparison of the RMSE values for speed and space headway prediction, respectively, between these models. The RMSE values of LSTM/RTS model indicate a superior performance compared to all other models. Note that the ARIMA model yielded a similar set of RMSE values compared to the LSTM model with CV penetration rates in excess of 50% (as with short-term traffic prediction). As indicated in Table III the RMSE and MAPE values using the LSM/RTS model combination for predicting speed and space headway exhibit superior performance with an increase in CV penetration.

## VII. REAL-TIME APPLICATION EFFICACY

The computational time required for a traffic flow parameters-prediction model must be very small for the real-time use of predicted traffic flow parameters in route planning and scheduling, assessing future traffic conditions, and for optimizing vehicle energy use. In our analyses, we aggregated individual vehicle-generated data at one-tenth of a second (100 milliseconds) time intervals to predict traffic speed and space headway. Given that real-time applications require instant aggregation of data analysis, we trained the LSTM model for a roadway corridor with a computation time within one-tenth of a second for real-time traffic prediction [41]. Our analyses results indicate that the LSTM/RTS model combination requires 70 milliseconds to predict speed and space headway, which is acceptable for real-time mobility and environmental applications [42]. We used an Intel(R) Core(TM) i5-3210M CPU@2.5GHz and 6.00GB installed memory to run the LSTM/RTS prediction model. We implemented the prediction model in a connected vehicle environment using a roadside unit (RSU) with a data processing unit (e.g.an Intel® NUC device) similar to that in our previous work [43]. We also used a Dedicated Short-Range Communication (DSRC) based RSU that communicates with connected vehicles (see Figure 15). We conducted a field experiment at the Clemson University-Connected and Autonomous Vehicle Testbed (CU-CAVT) to determine the two-way communication latency through DSRC between the CVs and a roadside unit for any real-time traffic application. In our field experiments on a roadway segment of Perimeter road at Clemson, South Carolina, we found that the two-way communication latency is 9 milliseconds on average (detailed in Table 3). However, given the variance in communication latency from trees, roadway slopes, and curvatures, we also determined maximum and minimum latencies of 22 and 5 milliseconds respectively for two-way communication between an RSU and a CV (again detailed in Table IV).

Our range of the total latency including computational time and two-way communication latency for our traffic flow parameters prediction application was 75 to 92 milliseconds. According to Southeast Michigan Test Bed Concept of Operations report, which is developed for the U.S. Department of Transportation (USDOT) to support connected vehicle research and development, the minimum latency for mobility and environmental applications should be within 1000 milliseconds [14]. As shown in Table 3, the total latency is much lower than the minimum latency requirement (approximately 1000 milliseconds), thus making our LSTM\RTS model suitable for real-time speed and space headway prediction.

## VIII. CONCLUSION

In this study, we detail a real-time prediction model that combines the Kalman filter based RTS noise reduction model with LSTM to improve prediction accuracy of the traffic flow parameters at low CV penetration rates. The average speed and average space headway data used in this study were generated from the Enhanced NGSIM dataset that contains vehicle trajectory data for every one-tenth of a second, which is similar to the broadcasting rate of BSMs in a CV environment.

We evaluated the prediction model to predict the average speed and average space headway using a CV penetration rate from 5% to 100%. The analyses revealed that this model was effective in predicting both speed and space headway for different penetrations of connected vehicles with no significant difference from the ground truth data. Compared to the baseline LSTM (without a noise reduction model), this combined LSTM/RTS model reduced the MAPE from 19% to 5% in terms of speed prediction; and from 27% to 9% in terms of space headway prediction, all at a 5% CV penetration. A comparison between LSTM/standard Kalman filter, LSTM/RTS filter and LSTM/moving average filters suggests that this LSTM/RTS combination achieves the best prediction performance in terms of RMSE and MAPE. The statistical significance test with a 95% confidence interval confirmed no significant difference between the predicted speed and space headway using this LSTM/RTS combination from the ground truth for all CV penetration levels. Moreover, the LSTM/RTS model outperforms the popular variation of RNN and statistical models in term of accuracy. Overall, the prediction accuracy of the average speed and space headway improves with an increase in CV penetration. The LSTM/RTS combination requires an average of 79 milliseconds to predict the traffic speed and space headway. This value is well within the bounds for real-time traffic prediction time requirement in a connected vehicle environment.

## ACKNOWLEDGMENTS

This material is based on a study supported by the Center for Connected Multimodal Mobility ($C^2M^2$) (USDOT Tier 1 University Transportation Center) Grant headquartered at Clemson University, Clemson, South Carolina, USA. Any opinions, findings, and conclusions or recommendations expressed in this material are those of the author(s) and do not necessarily reflect the views of the Center for Connected Multimodal Mobility ($C^2M^2$), and the U.S. Government assumes no liability for the contents or use thereof.

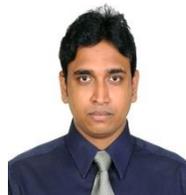

**Mizanur Rahman** received his Ph.D. and M.Sc. degree in civil engineering with transportation systems major in 2018 and 2013, respectively, from Clemson University. Since 2018, he has been a research associate of the Center for Connected Multimodal Mobility ($C^2M^2$), a U.S. Department of Transportation Tier 1 University Transportation Center (cecas.clemson.edu/c2m2) at Clemson University. He was closely involved in the development of Clemson University Connected and Autonomous Vehicle Testbed (CU-CAVT). His research focuses on transportation cyber-physical systems for connected and autonomous vehicles and for smart cities.

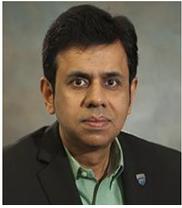

**Mashrur Chowdhury** (SM'12) received the Ph.D. degree in civil engineering from the University of Virginia, USA in 1995. Prior to entering academia in August 2000, he was a Senior ITS Systems Engineer with Iteris Inc. and a Senior Engineer with Bellomo–McGee Inc., where he served as a Consultant to many state and local agencies, and the U.S. Department of Transportation on ITS related projects. He is the Eugene Douglas Mays Professor of Transportation with the Glenn Department of Civil Engineering, Clemson University, SC, USA. He is also a Professor of Automotive Engineering and a Professor of Computer Science at Clemson University. He is the Director of the USDOT Center for Connected Multimodal Mobility (a TIER 1 USDOT University Transportation Center). He is Co-Director of the Complex Systems, Data Analytics and Visualization Institute (CSAVI) at Clemson University. Dr. Chowdhury is the Roadway-Traffic Group lead in the Connected Vehicle Technology Consortium at Clemson University. He is also the Director of the Transportation Cyber-Physical Systems Laboratory at Clemson University. Dr. Chowdhury is a Registered Professional Engineer in Ohio, USA. He serves as an Associate Editor for the IEEE TRANSACTIONS ON INTELLIGENT TRANSPORTATION SYSTEMS and Journal of Intelligent Transportation Systems. He is a Fellow of the American Society of Civil Engineers and a Senior Member of IEEE.

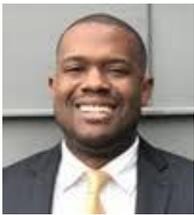

**Jerome L. McClendon** is a Research Assistant Professor at Clemson University. His appointment is currently in the Department of Automotive Engineering, but in the past he was a Research Assistant Professor in the School of Computing at Clemson University. Dr. Jerome McClendon's research efforts lie at the intersection of Artificial Intelligence and Human Centered Computing. Specifically, his research is focused on the design, implementation and evaluation of intelligent computing systems that mimic human behavior.